\documentclass{article}
\usepackage{ijcai23}

\usepackage{times}
\usepackage{soul}
\usepackage{url}
\usepackage[hidelinks]{hyperref}
\usepackage[utf8]{inputenc}
\usepackage[small]{caption}
\usepackage{graphicx}
\usepackage{amsmath}
\usepackage{amsthm}
\usepackage{booktabs}
\usepackage{algorithm}
\usepackage{algorithmic}
\usepackage[switch]{lineno}

\usepackage{booktabs}
\usepackage{url}
\usepackage{array}
\usepackage{subfigure}
\usepackage{multicol}
\usepackage{amssymb}
\usepackage{color}

\newcommand{\etal}{{\em et al.}}

\newtheorem{example}{Example}

\title{COOL, a Context Outlooker, and its Application
to Question Answering\\ and other Natural Language Processing Tasks}

\author{
Fangyi Zhu$^1$
\and
See-Kiong Ng$^1$\And
Stéphane Bressan$^1$
\affiliations
$^1$National University of Singapore\\
\emails
zhufy1030@gmail.com,\{seekiong, steph\}@nus.edu.sg
}

\begin{document}

\maketitle

\begin{abstract}
Vision outlooker improves the performance of vision transformers, which implements a self-attention mechanism by adding an outlook attention, a form of local attention.  

In natural language processing, as has been the case in computer vision and other domains, transformer-based models constitute the state-of-the-art for most processing tasks. In this domain, too, many authors have argued and demonstrated the importance of local context.

We present an outlook attention mechanism, COOL, for natural language processing. COOL, added on top of the self-attention layers of a transformer-based model, encodes local syntactic context considering word proximity and more pair-wise constraints than dynamic convolution used by existing approaches.

A comparative empirical performance evaluation of an implementation of COOL with different transformer-based models confirms the opportunity for improvement over a baseline using the original models alone for various natural language processing tasks, including question answering. The proposed approach achieves competitive performance with existing state-of-the-art methods on some tasks.
\end{abstract}

\section{Introduction}
Transformer neural networks, based solely on the self-attention mechanism~\cite{vaswani_attention_2017}, have been shown to be remarkably successful in handling various tasks in natural language processing (NLP)~\cite{lan2019albert,He2021DeBERTaDB} and other domains~\cite{vit,Sun2019BERT4RecSR,dong2018speech}.
Self-attention, in a nutshell, considers every token in the input sequence when encoding a current token. This strategy allows the learning and capture of cross-passage long-range dependencies. 

The interrogation naturally arises of whether self-attention is detrimental to local information and of how, if the case, this flaw can be remedied.
Several authors have considered and evaluated various ways of incorporating local context information back into a self-attention mechanism for various tasks in various domains~\cite{wu2019pay,gulati20_interspeech,volo,early_conv}. 

\begin{figure}[!t]
\centering
\includegraphics[width=\columnwidth]{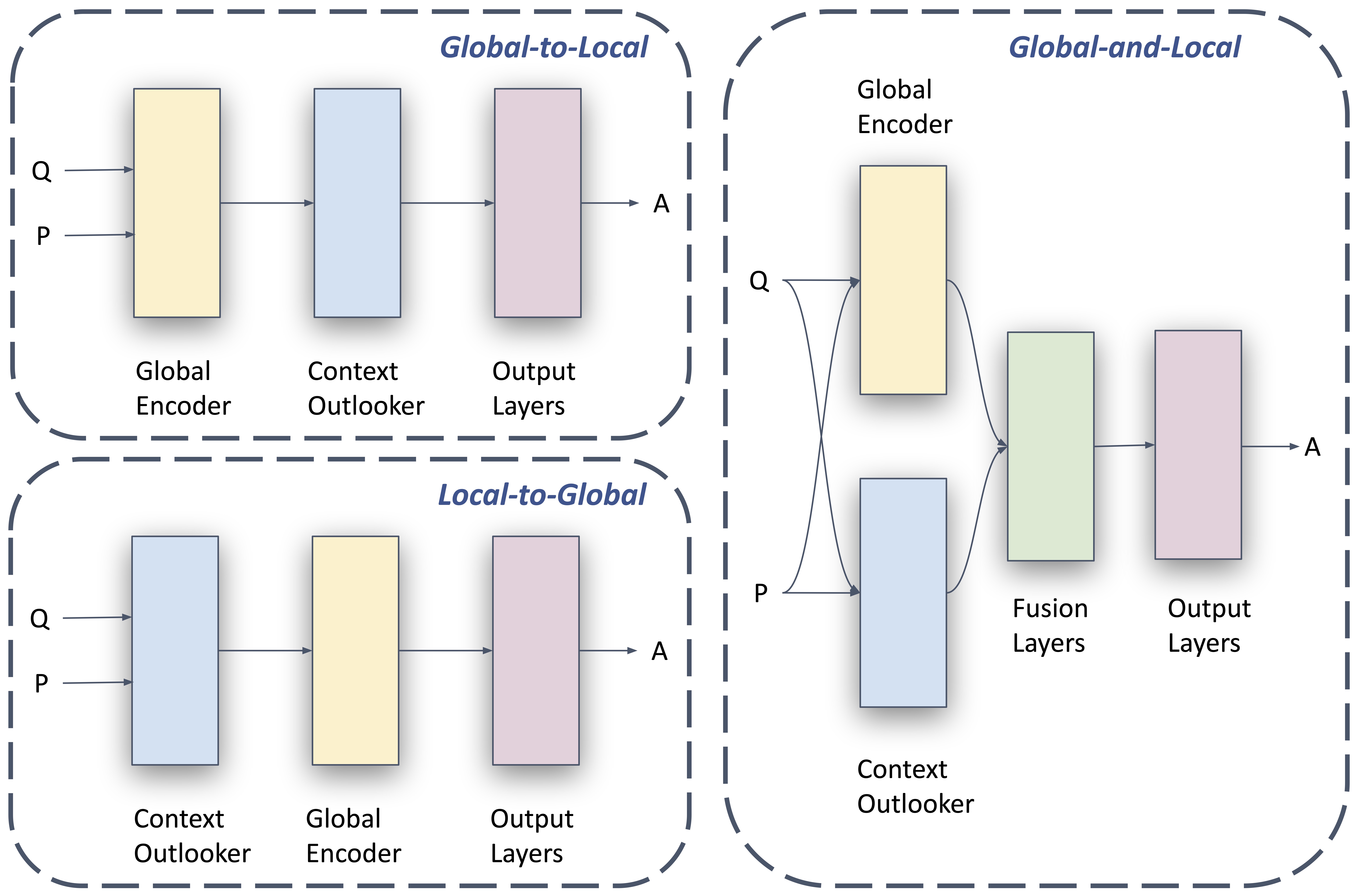} 
\caption{Different modes of integration of global and local encoding.}
\label{fig:more_structures}
\end{figure}

Recently, Yuan~\etal~\shortcite{volo} proposed Vision Outlooker (VOLO) for visual recognition. VOLO successfully improves the performance of vision transformers by adding a light-weight outlook attention mechanism, a form of local attention. 

We present and evaluate the outlook attention mechanism for NLP, named  
\textbf{Co}ntext \textbf{O}ut\textbf{l}ooker (COOL).
COOL implements a context outlook attention mechanism that encodes the local syntactic context considering word proximity and can be inserted into most transformer-based models. Compared to the operations for local augmentation used by existing works, outlook attention considers more pair-wise constraints as it processes multiple local windows containing an anchor token.
We consider three possible modes for the composition of global and local encoding, in which the global encoding precedes, \emph{Global-to-Local}, follows, \emph{Local-to-Global}, or is parallel to, \emph{Global-and-Local}, the local encoding.  

Comparative empirical performance evaluation of COOL with different transformer-based approaches confirms the opportunity for improvement over a baseline using the original models for eleven NLP tasks, including question answering, question generation, sentiment analysis, etc. 

\section{Related Work}

\paragraph{Transformer-based language models.}
Vaswani~\etal~\shortcite{vaswani_attention_2017} first proposed the Transformer architecture, based solely on an attention mechanism, for neural machine translation (NMT). Since Bidirectional Encoder Representation from Transformer (BERT)~\cite{devlin-etal-2019-bert} outstandingly tackled eleven NLP tasks, fine-tuning a transformer-based language model has become the mainstream approach for most NLP tasks. The transformer-based structure has been explored for other domains as well and achieved competitive performance with classical convolutional neural network- (CNN-) or recurrent neural network- (RNN-) based 
models, such as Vision Transformer (ViT) for 
computer vision (CV)~\cite{vit}, Bidirectional Encoder Representations from Transformers for Sequential Recommendation (BERT4Rec) for recommendation system~\cite{Sun2019BERT4RecSR}, and Speech-Transformer for automatic speech recognition (ASR)~\cite{dong2018speech}.

The self-attention mechanism, an operation for cross-passage encoding derived from ~\cite{cheng2016long}, is the crucial module in a transformer-based architecture. It relates all the elements in a sequence to capture long-range dependencies~\cite{nmt2014}. 
Nonetheless, there are two drawbacks of long-range encoding: the computation is expensive, especially for long sequences, and the cross-passage encoding might dilute the local information within the surrounding context.
To address these issues, recent works from different domains proposed to exploit an adaptive span~\cite{local_for_long_text,yang-etal-2018-modeling,sukhbaatar-etal-2019-adaptive}, or augment local encoding, e.g., ~\cite{wu2019pay,zhang2020dynamic,jiang2020convbert} for NLP,~\cite{swin,volo} for CV, and~\cite{gulati20_interspeech} for ASR.

\paragraph{Local attention.}
The dilution of local information in a standard self-attention mechanism for NLP and the benefit of its reinstatement was first discussed for NMT by Wu~\etal~\shortcite{wu2019pay}. They presented the dynamic convolution to replace global encoding with the self-attention layers. They demonstrated that only encoding the local context could outperform executing a global encoding via self-attention. 

Several works followed that tried to integrate both local, leveraging the dynamic convolution technique, and global encoding, leveraging the self-attention mechanism. Zhang~\etal~\shortcite{zhang2020dynamic} presented a novel module where a dynamic convolution and a self-attention mechanism were placed in parallel for NMT. Jiang~\etal~\shortcite{jiang2020convbert} further connected a self-attention mechanism and a dynamic convolution in series to form a mixed-attention mechanism.

Recently, in the area of CV, Yuan~\etal~\shortcite{volo} and Xiao~\etal~\shortcite{early_conv} demonstrated that a significant factor limiting the performance of the transformer-based fashion for visual tasks was the low efficacy of the self-attention mechanism in encoding the local features as well. Moreover, Yuan~\etal~\shortcite{volo} proposed VOLO for efficient pixel token aggregation via an outlook attention mechanism, with which a transformer-based model could outperform the classical CNN-based models on visual recognition and segmentation. 
Compared to the standard dynamic convolution technique, handling a single constraint within the current window only, the outlook attention mechanism deals with more pair-wise constraints for the current token encoding. 
Specifically, it considers all local windows containing the current token.

\section{Methodology}

We present COOL, an outlook attention mechanism and architecture, that helps emphasise local information in NLP using a transformer-based model.

For the sake of clarity, we present the proposed model for a specific task of extractive question answering.

\subsection{Architecture}
\label{section:structures-overview}

\begin{figure*}[!t]
\centering
\includegraphics[width=\textwidth]{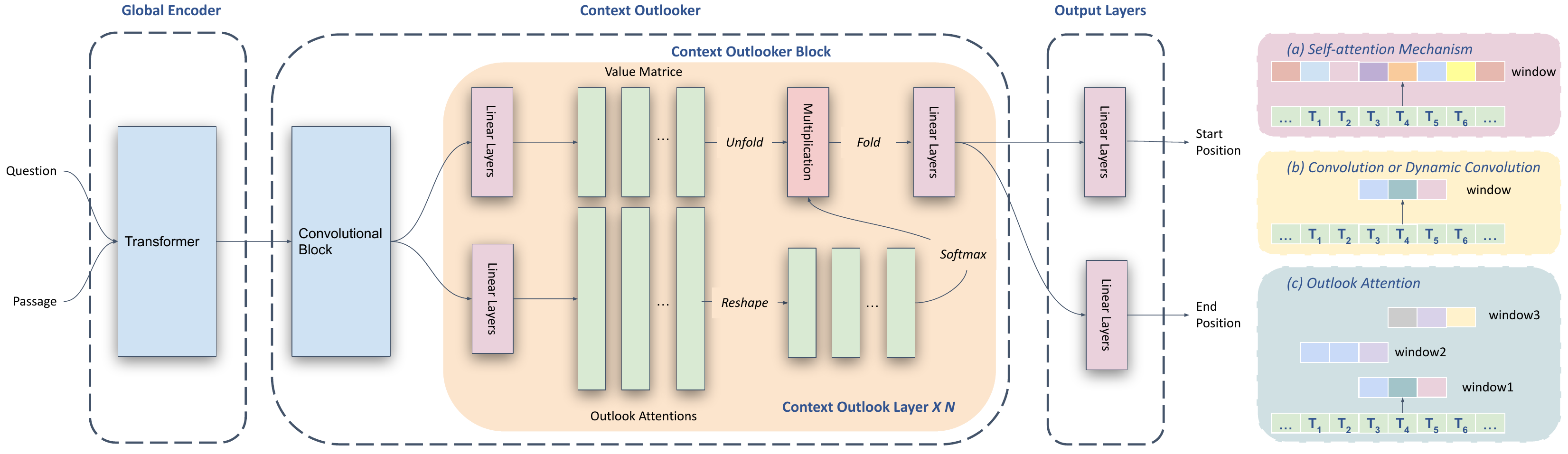}
\caption{Diagram of our proposed system with a \textit{G2L} structure. There are three major components: first, global encoding by a transformer-based encoder; second, local augmentation by the context outlooker; third, answer prediction by the output layers. \textit{Unfold} and \textit{Fold} correspond to unfold and fold operations in PyTorch.}
\label{fig:diagram}
\vspace{-4mm}
\end{figure*}

Yuan~\etal~\shortcite{volo} argued the importance of local information for visual tasks and proposed VOLO to overcome the weakening of local information in representations produced by the standard self-attention mechanism. The outlook attention mechanism, the primary component of VOLO, dynamically fine-tunes pixel token features produced by the self-attention mechanism considering local information from surrounding pixel tokens. 

We present to generalise this idea into NLP as the context outlooker (COOL).
The overall architecture has three major modules, global encoding by a generic transformer-based encoder, local encoding via the proposed context outlooker, and output layers for the downstream tasks. The output layers for an extractive question answering task are for answer prediction.

There are three different modes for integrating global and local encoding illustrated in Figure~\ref{fig:more_structures},
\begin{itemize}
    \item \emph{Global-to-Local (G2L)}: first encodes the question and passage by the global encoder. Then, the yielded global embeddings go through the context outlooker to enrich the local context information. Finally, the start and end positions of the answer are predicted in the output layers. 
    \item \emph{Local-to-Global (L2G)}: switches the order of global encoding and local augmentation in the first structure. In contrast to \emph{G2L}, \emph{L2G} first conducts local encoding, followed by global encoding.
    \item \emph{Global-and-Local  (G\&L)}: executes global and local encoding parallelly and fuses the integrated global and local embeddings with several linear layers. Finally, predict the answer locations.
\end{itemize}

Next, we elaborate the workflow in light of the first structure of \emph{G2L} in Figure~\ref{fig:diagram}.

\subsection{Global Encoder}

Given a question $q = \left \{ q^{1},...,q^{M} \right\}$ and a passage $p = \left \{ p^{1},...,p^{N} \right\}$,  $M$, $N$ denotes the total number of tokens in the question and passage. 
The global encoder, as a backbone of our proposed system, can be the encoder in any transformer-based model, e.g., BERT~\cite{devlin-etal-2019-bert}, ALBERT~\cite{lan2019albert}, and RoBERTa~\cite{zhuang-etal-2021-robustly} with an encoder-only architecture, or ProphetNet~\cite{qi-etal-2020-prophetnet} with an encoder-decoder architecture.
\begin{equation}
    h_{g} = Global(\left[ q;p \right])),
\end{equation}
where $h_{g} \in \mathbb{R}^{L \times H} $ is the obtained global representations, $L$ is the max length of the input sequence, and $H$ is the size of hidden units in the last layer of the global encoder.

\subsection{Context Outlooker}

Context outlooker, the pivotal component in our proposed approach, consists of a convolutional block and a context outlook block.


\footnotetext{Unfold extracts sliding local blocks from a batched input tensor. Fold, the reverse operation of unfold, combines an array of sliding local blocks into the original large containing tensor.}

\subsubsection{Convolutional Block}
The convolution operation is effective and stable for capturing local information~\cite{kalchbrenner-etal-2014-convolutional}. Several works have proved that inserting convolutional layers can help transformers in applications to different domains~\cite{wu2019pay,early_conv,karpov2020transformer,liu2021transformer,gulati20_interspeech}. 
We thus try to insert a convolutional block for local augment, and further compare the convolution technique to our proposed module.


We stack the convolutional block, consisting of several convolutional layers, on top of the global encoder,
\begin{align}
     \label{eq:get_hc} h_{c}&=ReLU(ConvBlock(h_{g})),\\
     \label{eq:hc}
     h_{c}&= \left [ h_{c}^{1},...,h_{c}^{d},...h_{c}^{D} \right ],\\
     \label{eq:conv_concat} \hat{h_{c}}&=Concat(AdaptivePool(h_{c})).
\end{align}
In Eq.~\eqref{eq:hc}, $h_{c}^{d}$ is the output of the $d$-th convolutional layer with ReLU non-linear activation operation~\cite{pmlr-v15-glorot11a}, and $D$ is the number of convolutional layers in the block. The dimension of $h_{c}^{d}$ is $L^{d} \times F^{d}$. $L^{d}$ depends on the kernel size and padding size of the convolutional layer, and $F^{d}$ is the number of filters in the convolutional layer. To align different $L^{d}$ to original length $L$, we perform adaptive pooling after each convolutional layer interpreted in Eq.\eqref{eq:conv_concat}. Finally, the outputs from all convolutional layers are concatenated along the second dimension to produce $\hat{h_{c}} \in \mathbb{R}^{L \times F}$, 
\begin{equation}
    F = \sum_{\quad 1 \leq d <D} F^{d}.
\end{equation}

\subsubsection{Context Outlook Block} 

The context outlook block, consisting of several context outlook layers, follows by the convolutional block. 
 
In analogy with the feature map of an image, each row in the representation matrix $\hat{h_{c}} $ is the representation of a token. To maintain the integrality of representation of each token, we adapt the local window size in the visual outlook attention mechanism to $K \times F$ to handle text tokens instead of $K \times K$ for processing image pixel tokens.

Similar to the computation of \emph{Query}, \emph{Key}, \emph{Value} matrices in the self-attention mechanism, the outlook attention mechanism first produces value matrix $v \in \mathbb{R}^{L \times F} $ via a linear layer with $\hat{h_{c}}$ from previous convolutional block,
\begin{align}
    \label{eq:v_vector} v &= W_{v}\hat{h_{c}} + b_{v},
\end{align}
where $W_{v}$ and $b_{v}$ separately denote the parameters and
bias needed to be learned in the linear layer.

Hence, for the $i$-th token, the value matrices of its neighbour tokens in a local window with the size of $K \times F$ are,
\begin{equation}
    \label{eq:unfold}
    v_{\Delta_{i}} = v_{i+r-\left\lfloor\frac{K}{2}\right\rfloor},\quad 0 \leq r <K,
\end{equation}
$v_{\Delta_{i}} \in \mathbb{R}^{K \times F} $, and $\lfloor\cdot\rfloor$ is the round down formula.



In visual outlook attention, Yuan~\etal~\shortcite{volo} proposed to produce the aggregating weights directly from the features of the centric token to dispense the expensive dot product computation in self-attention, as each spatial location in an image feature map can represent its close neighbours. 
Likewise, the embedding of a token in a sentence is able to express its neighbour tokens as well. Therefore, following their work, we calculate attention matrix $a \in \mathbb{R}^{L \times A}$ with $\hat{h_{c}}$ via another linear layer, 
\begin{equation}
   \label{eq:a_vector} a = W_{a}\hat{h_{c}} + b_{a},
\end{equation}
here $A=K\times K\times F$. $W_{a}$ and $b_{a}$ are learned parameters and bias. Then, the attention weights within different local windows are obtained by reshaping $a$ from $\mathbb{R}^{L \times A}$  to $\mathbb{R}^{L \times K \times (K \times F)}$ , $a_{\Delta_{i}} \in \mathbb{R}^{1 \times 1 \times (K \times F)}$ represents the obtained attention weights of $i$-th token within a local window.


Therewith, normalise the attention weights in each local window with a softmax function and aggregate the neighbours within the window with the acquired attention weights,
\begin{equation} 
    \hat{v}_{\Delta_{i}}= softmax(a_{\Delta_{i}})v_{\Delta_{i}}.
\end{equation}

Afterward, sum up the weighted values at the same position from different windows,
\begin{equation} 
    \label{eq:fold}
    \tilde{v}_{i}= \sum_{0 \leq r <K} \hat{v}_{\Delta_{i+r-\left\lfloor\frac{K}{2}\right\rfloor}}^{i},
\end{equation}
where $\tilde{v}_{i}$ is the acquired representation for $i$-th token with rich local context information, and $\tilde{v}$ is the representation matrice for the whole sequence.


Finally, conduct a residual operation~\cite{he2016residual} that sum the encoded $\tilde{v}$ with $\hat{h_{c}}$, and go through several linear layers for further representation learning, followed by performing another residual operation,
\begin{align}
    h_{f} &= \tilde{v} + \hat{h_{c}},\\
    \hat{h}_{f} &= W_{f}h_{f}+b_{f},\\
    h^{'}_{f} &= \hat{h}_{f}+h_{f}.
\end{align}

\subsection{Output Layers}

We stack linear layers to perform answer prediction with the obtained $h^{'}_{f}$,
\begin{align}
    y_{s} &= \text{softmax}(W_{s}h^{'}_{f}+b_{s}),\\
    y_{e} &= \text{softmax}(W_{e}h^{'}_{f}+b_{e}),
\end{align}
$W_{s}$, $W_{e}$ and $b_{s}$, $b_{e}$ are parameters and bias needed to be learned in the linear layers. $y_{s}, y_{e}$ are the probability distributions for the start position and the end position.

The training objective is to minimize the sum of the negative log probabilities at the ground truth start position $a_{s}$ and end position $a_{e}$,
\begin{equation}
\mathcal{L}=-\log (y_{s}^{a_{s}}+y_{e}^{a_{e}}).
\end{equation}

\section{Experiments}
\label{sec:perf}
We evaluate the proposed system for extractive question answering on the Stanford Question Answering Dataset (SQuAD), i.e., SQuAD 2.0~\cite{rajpurkar-etal-2018-know} and SQuAD 1.1~\cite{rajpurkar-etal-2016-squad}, with no additional training data. We use the standard metrics for question answering, F1 measure and exact match (EM).

We conduct extensive experiments with other NLP tasks and other language to evaluate our proposed approach's versatility and significance. 

\subsection{Setup}
We leverage the encoder in BERT-base\footnote{\textit{bert-base-cased}: \url{https://huggingface.co/bert-base-cased}}, ALBERT-base\footnote{\textit{albert-base-v2}: \url{https://huggingface.co/albert-base-v2}}, RoBERTa-base\footnote{\textit{roberta-base-squad2}: \url{https://huggingface.co/roberta-base}}, and ProphetNet\footnote{https://github.com/microsoft/ProphetNet} for text generation as the global encoder and keep their default settings. The max sequence length in our experiments is set to 384.

In the convolutional block, we stack multiple convolutional layers to augment local information within different n-grams. We stack $3 \times (H+4)$, $4 \times (H+4)$, and $5 \times (H+4)$ convolutional layers with 100 filters each. $H$ is the dimension of the input representations, for instance, $H$ is 768 while using BERT-base as global encoder. 2-padding is used to maintain the sequence length.

In the context outlook block, the kernel size of context outlook layers is $3 \times 300$, depending on the number of filters in the convolutional layers. While without a convolutional block, it is set to $3 \times H$ according to the global encoder. 

During training, we use AdamW~\cite{Loshchilov2019DecoupledWD} optimizer. We set a learning rate to 2e-5 for the global encoder, and 1e-4 for other layers. We train the model with a mini-batch size of 48 for about 5 epochs. The main codes are uploaded as supplementary materials. 

\subsection{Extractive Question Answering}

\paragraph{Choice of architecture.}
We first compare the three modes of global and local encoding integration: {\em G2L}, {\em L2G}, and {\em G\&L}.

\begin{table}[!t]
\centering
\small
    \begin{tabular}{lccc}
    \toprule
    Architecture & Outlook Layers & F1 & EM \\
    \midrule
    BERT & - & 75.41 & 71.78 \\
    \midrule
    {\em G2L} w/o Conv.  & 3 & 76.79 & 73.36 \\
    {\em G2L} w/ Conv.   & 2 & {\bf 77.11} & {\bf 74.02 }\\
    \midrule
    {\em L2G} w/o Conv.  & 3 & 74.80 & 72.02 \\
    {\em L2G} w/ Conv.   & 2 & 50.06 & 49.98 \\
    \midrule
    {\em G\&L} w/o Conv.  & 3 & 75.58 & 72.78 \\
    {\em G\&L} w/ Conv.   & 2 & 75.98 & 72.86 \\
    \bottomrule
    \end{tabular}
\caption{SQuAD 2.0 Dev results of different integration modes.}
\vspace{-5mm}
\label{tab:comparisons_of_different_structures}
\end{table}

Table~\ref{tab:comparisons_of_different_structures} reports the results of the respective architectures with BERT as a baseline. 
The table shows the most effective mode is {\em G2L}.
{\em G\&L} achieves a minor improvement over the baseline, yet not exceeding that of the first structure. A possible reason is it's difficult to precisely capture the relationships between global and local representations.
We find {\em L2G} architecture does not work well. The local encoding operation perturbs the following global encoding procedure, particularly while integrating the convolution operations with different kernel sizes. Specifically, directly combining the information within different n-grams confuses to the model.
In summary, it should better not replace the input to the pre-trained model. We nevertheless acknowledge that early convolution might benefit a scenario as discussed in~\cite{early_conv} for visual tasks.
In the following experiments, we adopt the {\em G2L} mode.

\begin{table}[!t]
\centering
\small
    \begin{tabular}{lcc}
    \toprule
    Model      & F1    & EM    \\
    \midrule
    {\em SQuAD 2.0}\\
    \midrule
    BERT                      & 75.41 & 71.78 \\
    COOL(BERT) w/o Conv.   & 76.79 & 73.36 \\
    COOL(BERT) w/ Conv.           & {\bf 77.11} & {\bf 74.02} \\
    \midrule
    ALBERT                    & 78.63 & 75.33 \\ 
    COOL(ALBERT) w/o Conv.   & 79.23 & 75.95 \\
    COOL(ALBERT) w/ Conv.            & {\bf 79.49} & \textbf{76.21} \\
    \midrule
    RoBERTa           & 82.91 & 79.87 \\
    COOL(RoBERTa) w/o Conv. & 83.39 & 80.12 \\
    COOL(RoBERTa) w/ Conv. & {\bf 83.61} & {\bf 80.33}\\
    \toprule
    {\em SQuAD 1.1}\\
    \midrule
    BERT                      & 88.33 & 80.76  \\
    COOL(BERT) w/o Conv.   & 88.38 & {\bf 81.00} \\
    COOL(BERT) w/ Conv.    & {\bf 88.44} & 80.91\\
    \midrule
    ALBERT                      & 87.81 & 79.50  \\
    COOL(ALBERT) w/o Conv.    & 87.81 & 79.93\\
    COOL(ALBERT) w/ Conv.     & {\bf 87.90} & {\bf 80.15}\\
    \midrule
    RoBERTa           & 90.40 & 83.00  \\
    COOL(RoBERTa) w/o Conv. & 90.65 & 83.36 \\
    COOL(RoBERTa) w/ Conv. & {\bf 90.66} & {\bf 83.42}\\
    \bottomrule
    \end{tabular}
\caption{SQuAD Dev results for extractive question answering.}
\label{tab:squad_results}
\vspace{-6mm}
\end{table}

\paragraph{Overall results.}
Table~\ref{tab:squad_results} shows the comparisons of COOL with and without the convolutional block combined with BERT, ALBERT, RoBERTa and the corresponding baselines for SQuAD 2.0 and SQuAD 1.1. 
It can be viewed that COOL with BERT achieves an improvement of 2.24 and 1.7 points for EM and F1, respectively, over the original BERT. COOL with ALBERT adds 0.88 EM points and 0.86 F1 points. COOL with RoBERTa outperforms the original in terms of both EM and F1 as well. 
SQuAD 1.1 does not contain unanswerable questions and is thus less challenging than SQuAD 2.0. Nevertheless, models inserted COOL improve across the board over the original models. These results confirm the benefit of an appropriate consideration of local information in a transformer-based approach for question answering that COOL is an effective approach.

The comparison of the results for COOL with and without the convolutional block, further shows that COOL with the convolutional block generally outperforms COOL without the convolution block, except for the case of SQuAD 1.1 where COOL(BERT) without the convolutional block marginally outperforms COOL(BERT) with the convolutional block by 0.09 point.
In the following ablation study, we further explore and compare the impacts of the convolution techniques and the outlook attention.

\begin{table}[!t]
\centering
\small
    \begin{tabular}{lcc} \\
    \toprule
    Model    & F1    & EM    \\
    \midrule
    BERT   & 75.41 & 71.78  \\
    BERT-Deeper(3 layers) & 75.28 & 71.81\\ 
    BERT-Deeper(5 layers) & 74.94 & 71.28\\
    COOL(BERT) w/o Conv. & 76.79 & 73.36 \\
    COOL(BERT) w/ Conv. & {\bf 77.11} & {\bf 74.02}\\
    \bottomrule
    \end{tabular}%
\caption{Comparisons with deeper models.}
\label{tab:deeperbase}
\end{table}

\begin{table}[!t]
\vspace{-6mm}
\centering
\small
    \begin{tabular}{lcc} \\
    \toprule
    Model    & F1    & EM    \\
    \midrule
    BERT & 75.41 & 71.78  \\
    BERT + Convolution & 75.82 & 72.81\\ 
    BERT + Dynamic Convolution & 75.57 & 71.19\\
    BERT + Visual Outlook Attention & 75.89 & 72.94 \\
    BERT + Context Outlook Attention & {\bf 76.79} & {\bf 73.36} \\
    \bottomrule
    \end{tabular}%
\caption{Comparisons with other techniques for local augmentation.}
\vspace{-4mm}
\label{tab:visualandcontext}
\end{table}

\paragraph{Ablation study.}

To verify these improvements are contributed by COOL rather than more trainable parameters, we first compare the performance with a deeper BERT containing more self-attention layers in Table~\ref{tab:deeperbase}. We added 3 or 5 more self-attention layers on top of the original BERT. It can be observed that direct stacking layers does not gain an improvement over the base model, even bringing adverse effects. It further confirms that COOL 
brings these improvements and it can remedy the deficiencies of the self-attention mechanism while encoding the local information.

We further compare the outlook attention with other techniques for local information enhancement in Table~\ref{tab:visualandcontext}. We separately implement BERT with standard convolution and dynamic convolution~\cite{wu2019pay,jiang2020convbert}. The table reports the COOL(BERT) results without the convolutional block in the interest of fairness. We can see that the models with outlook attention outperform those with standard or dynamic convolution.

We also compare the results between the original visual outlook attention and context outlook attention to demonstrate that the changes to the visual outlook attention to applying to NLP are relevant and effective.
The results are collated in Table~\ref{tab:visualandcontext} as well.
It can be viewed that outlook attention, whether implemented as the visual outlook or improved into a mechanism dedicated to NLP, brings an improvement. 
Thus, the modified module is indeed better suited to NLP tasks.

\begin{figure}[!t]
    \centering
    \includegraphics[width=0.7\columnwidth]{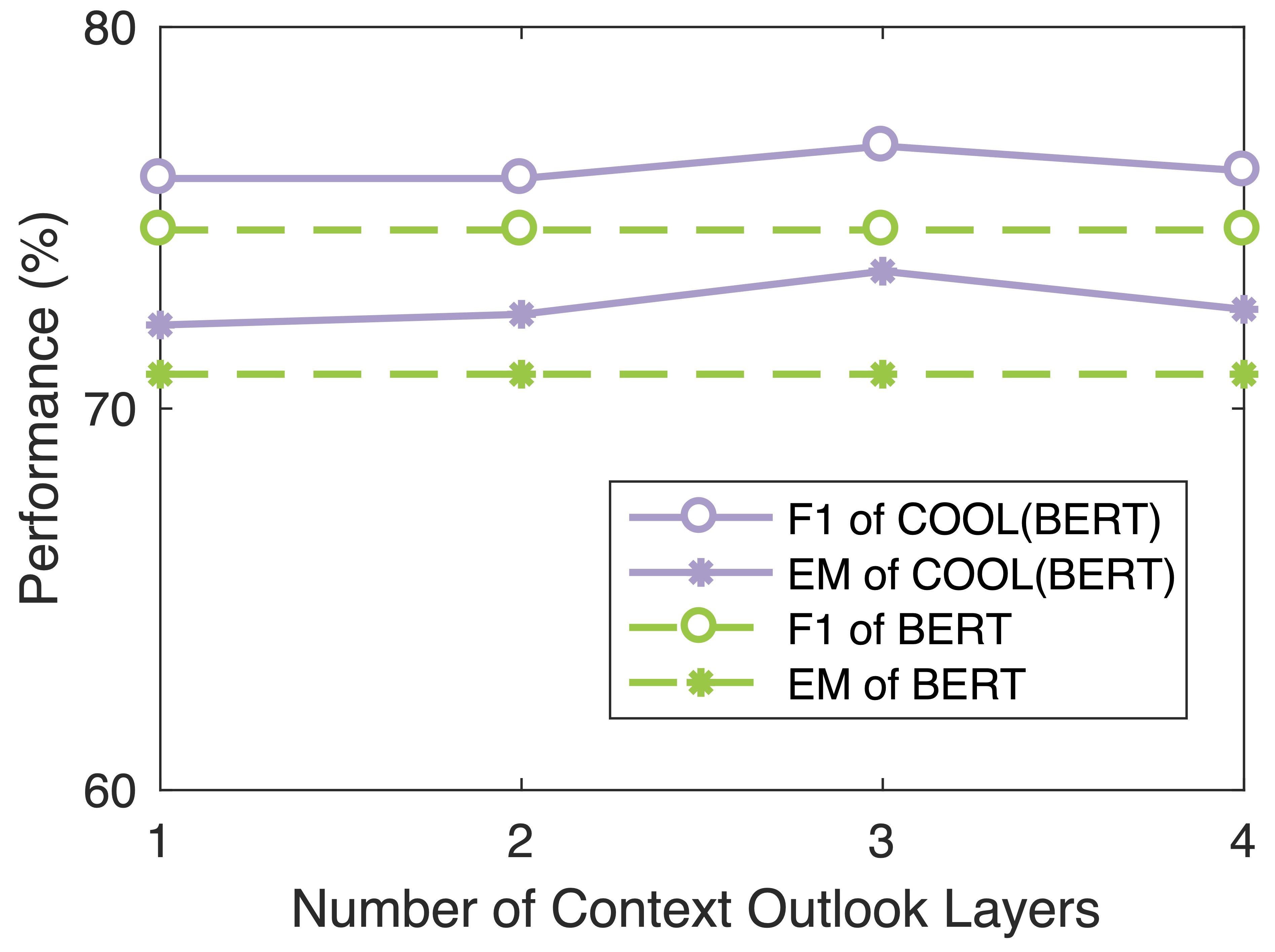} 
    \caption{Results with different numbers of context outlook layers.}
    \label{fig:noconv}
    \vspace{-4mm}
\end{figure}

The results of COOL with different numbers of context outlook layers are reported in Figure~\ref{fig:noconv}. All variants with COOL outperform the baseline. 
It achieves peak performance with two context outlook layers with a convolutional block. 

\paragraph{Illustrative examples.}

We first present and discuss illustrative examples of the situations in which models with COOL predict the correct or better answers while the baseline models do not and reverse.
The examples indicate the question (Q), the paragraph or the significant excerpt thereof (P), the answers by the baseline models, and the answers by COOL with the corresponding baseline models. The parts of the paragraph relevant to the discussion are highlighted in blue and correct answers in green.

\begin{figure}[!t]
\vspace{-7mm}
\centering
\subfigure[BERT]{
\includegraphics[width=0.47\columnwidth]{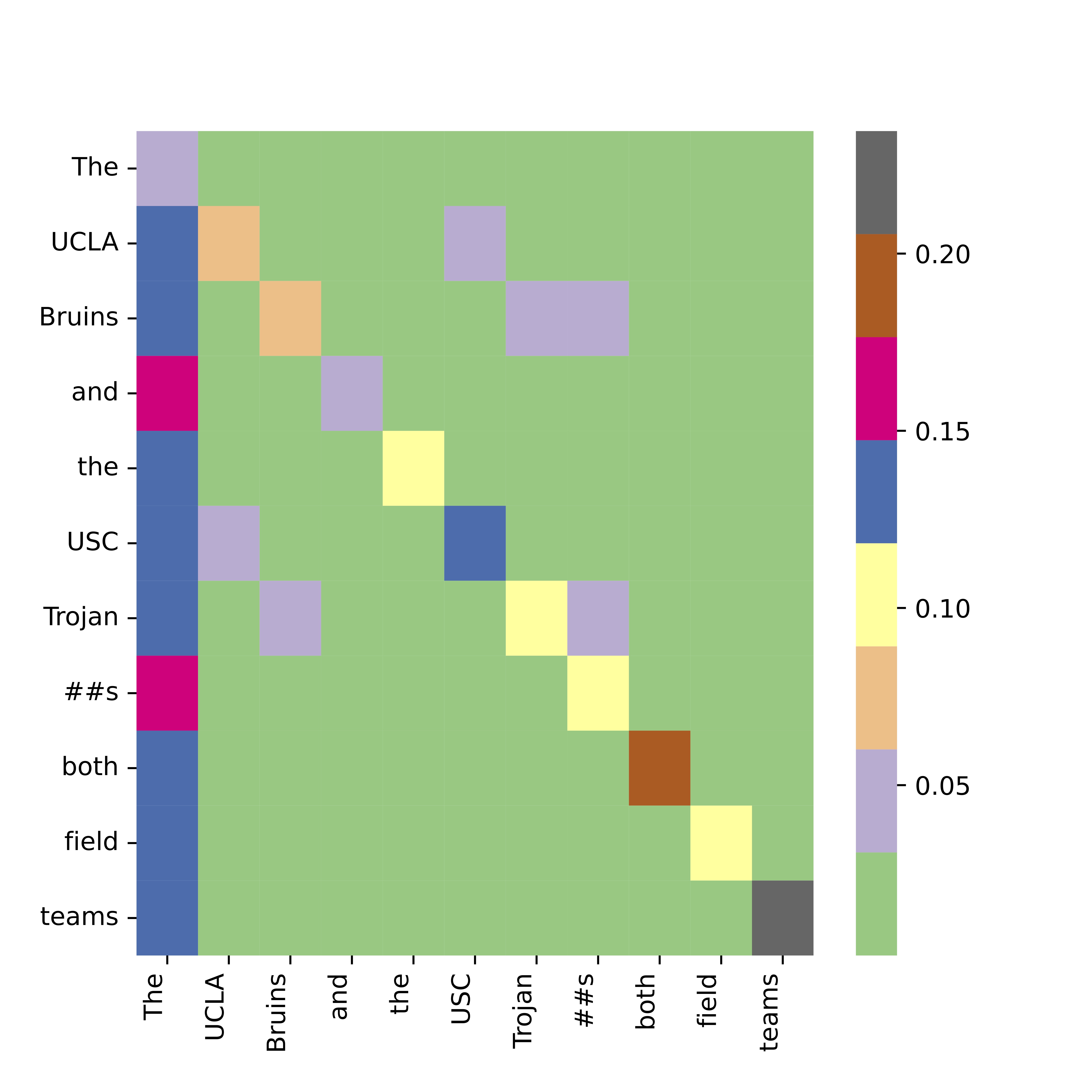}
}
\subfigure[RoBERTa]{
\includegraphics[width=0.47\columnwidth]{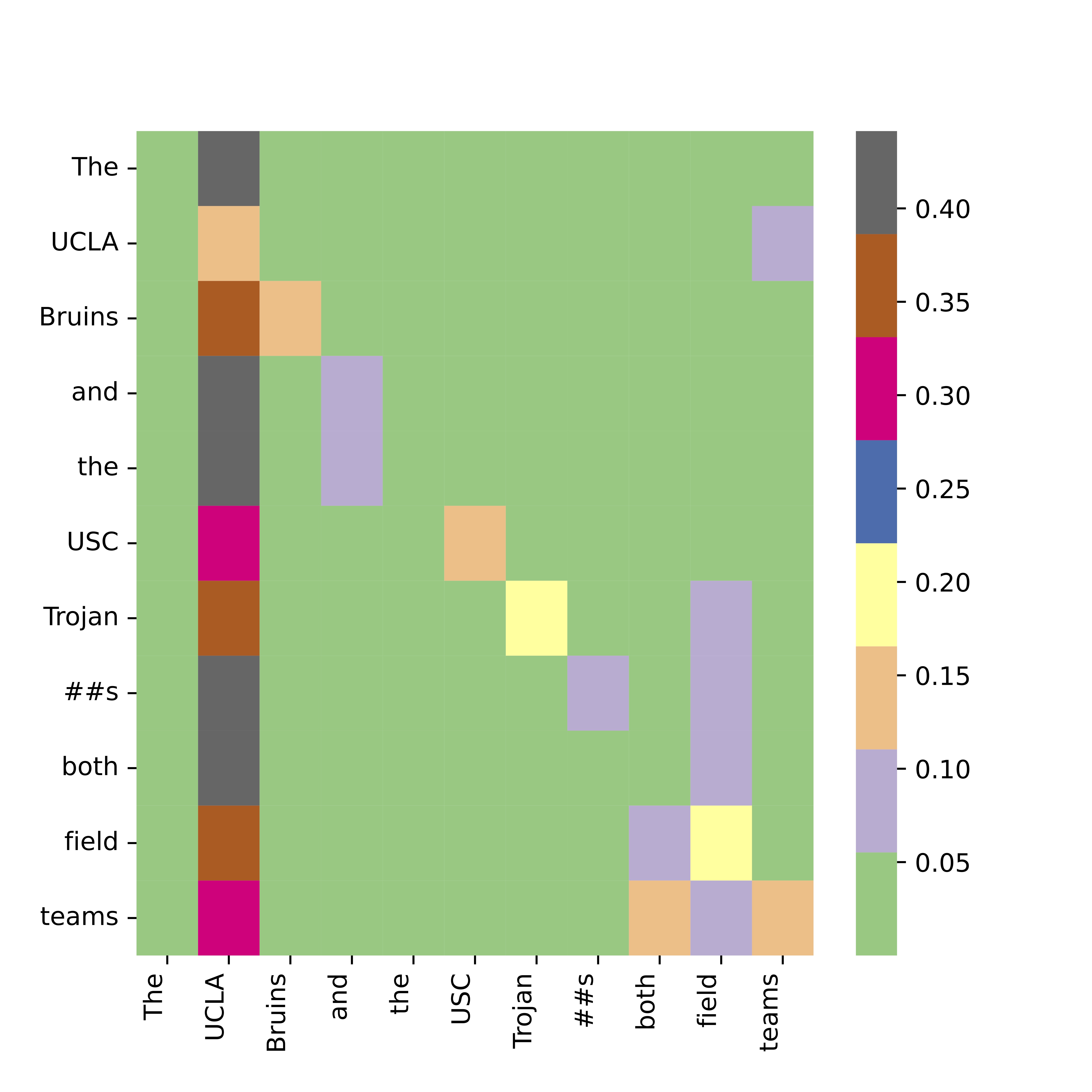}
}
\vspace{-3mm}
\caption{Visualization of weights yielded by the self-attention mechanism.}
\vspace{-4mm}
\label{fig:attention-heatmap}
\end{figure}

In Example~\ref{example11}, the baseline models fail to predict the correct answer. After inserting COOL, COOL(BERT) and COOL(RoBERTa) precisely recognise the relationship between `USC' and `Trojans' and thus give the correct answer. Although COOL(ALBERT) still fails, it attempts to recognise the relationship between the state `UCLA' and the team's name `Bruins'.  We further visualize the self-attention weights in Figure~\ref{fig:attention-heatmap}. 
We can observe the most important tokens to `USC' are itself, `UCLA', `both', in that order. For `UCLA', the important tokens are itself, `USC', `the', `Bruins', `and', and `both'. 
Likewise, the important words to `Bruins' are itself, `both', `and', `Trojan', and to `Trojan' are itself, its subtoken `\#\#s', and `Bruins'.
It explains that the standard self-attention operation via similarity score calculation can not distinguish these tokens and understand their relationships. It thus predicts a long span as the answer. 
Whereas augmenting the local context information by COOL helps to address this disadvantage and benefits understanding the relations between close neighbours.

\begin{small}
\begin{example}\hfill

\noindent Q: What is the {\color{blue} name of the team from USC}?


\noindent P: {\color{blue} The UCLA Bruins and the USC Trojans} both field teams...

\noindent\begin{minipage}[t]{0.48\columnwidth}

BERT: UCLA Bruins and the USC Trojans

ALBERT: UCLA Bruins 

RoBERTa: USC Trojans
\end{minipage}
\begin{minipage}[t]{0.48\columnwidth}

COOL(BERT):  {\color{green} Trojans}

COOL(ALBERT): Bruins

COOL(RoBERTa): {\color{green} Trojans}

\end{minipage}
\label{example11}
\end{example}
\end{small}

Moreover, we showcase interesting Example~\ref{example15}, the consideration of the local context allows COOL to capture logical relations where the baseline models fail. It is reasonable as most logical formulas are within local content.

\begin{small}
\begin{example}\hfill

\noindent Q: What is an expression that can be used to illustrate the suspected inequality of complexity classes?

\noindent P: Many known complexity classes are suspected to be unequal,... For instance {\color{blue} P $\subseteq$ NP $\subseteq$ PP $\subseteq$ PSPACE}, but it is possible that P = PSPACE. 

\noindent\begin{minipage}[t]{0.48\columnwidth}

BERT: \textlangle no answer\textrangle 

ALBERT: P = PSPACE 

RoBERTa: P = PSPACE 

\end{minipage}
\begin{minipage}[t]{0.48\columnwidth}

COOL(BERT): {\color{green} P $\subseteq$ NP $\subseteq$ PP $\subseteq$ PSPACE, but it is possible that P = PSPACE.}

COOL(ALBERT):  {\color{green} P $\subseteq$ NP $\subseteq$ PP $\subseteq$ PSPACE, }

COOL(RoBERTa):  {\color{green} P $\subseteq$ NP $\subseteq$ PP $\subseteq$ PSPACE}
\end{minipage}
\label{example15}
\end{example}
\end{small}

There are, unfortunately, situations in which too much local attention can be misleading and detrimental, as illustrated by Examples~\ref{example13}. We target to address this challenge of balancing the global and local encoding in future work.

\begin{small}
\begin{example}\hfill

\noindent Q: What was the Anglo-Norman language's final form? 

\noindent P: The  {\color{blue} Anglo-Norman} language was eventually absorbed into the Anglo-Saxon language of their subjects (see Old English) and influenced it, helping (along with the Norse language of the earlier Anglo-Norse settlers and the Latin used by the church) in the development of Middle English. It in turn evolved into {\color{blue} Modern English}.

\noindent\begin{minipage}[t]{0.48\columnwidth}

BERT: {\color{green} Modern English}

ALBERT: \textlangle no answer\textrangle 

RoBERTa: {\color{green} Modern English}
\end{minipage}
\begin{minipage}[t]{0.48\columnwidth}

COOL(BERT): \textlangle no answer\textrangle 

COOL(ALBERT): \textlangle no answer\textrangle 

COOL(RoBERTa): {\color{green} Modern English}
\end{minipage}
\label{example13}
\end{example}
\end{small}

\subsection{Other Natural Language Processing Tasks}


We seek to confirm the versatility of COOL and its effectiveness in tackling more NLP tasks. We compare the baseline model and a COOL-augmented version, noted COOL (\emph{\textlangle base\textrangle}), for question generation, multiple-choice question answering, natural language inference, sentiment analysis, named-entity recognition, and part-of-speech tagging with the corresponding widely used datasets.

\paragraph{Question generation.} 
Question generation (QG) targets generating a question given a passage and an answer~\cite{qi-etal-2020-prophetnet}.  We evaluate the model with SQuAD 1.1 and adopt the commonly used metrics, BLEU@4 (B@4)~\cite{papineni-etal-2002-bleu}, METEOR (M)~\cite{banerjee-lavie-2005-meteor}, and ROUGE-L (R)~\cite{lin-2004-rouge}.  
ERNIE-GEN~\cite{xiao2020ernie} and ProphetNet~\cite{qi-etal-2020-prophetnet} achieve the state-of-the-art for QG. 
We insert COOL into ProphetNet, then compare it with the original ProphetNet and with ERNIE-GEN. 
The results of the different methods are compared in Table~\ref{tab:qg_results}. The original ProphetNet has a BLEU@4 score of 23.91 and a METEOR score of 26.60.
With COOL, the BLEU@4 and METEOR reach 25.93 and 27.14, significantly outperforming both the original ProphetNET and ERNIE-GEN. 

\begin{table}
\centering
    \small
    \begin{tabular}{lccc} \toprule
    Model   & B@4  & M & R \\ 
    \midrule
    ProphetNet & 23.91 & 26.60 & 52.26\\
    ERNIE-GEN (beam=1)  & 24.03 & 26.31 & 52.36\\
    ERNIE-GEN (beam=5) & 25.40 & 26.92 & 52.84\\
    \midrule
    COOL(ProphetNet) (beam=1)  &24.33 & 26.24 & 52.53      \\
    COOL(ProphetNet) (beam=5)  &\textbf{25.93}& \textbf{27.14} & \textbf{52.86}\\  
    \bottomrule
    \end{tabular}%
\caption{SQuAD 1.1 Dev results for question generation.}
\label{tab:qg_results}
\end{table}

\begin{table}[!t]
\centering
    \resizebox{\columnwidth}{!}{
    \begin{tabular}{lcccccc}
    \toprule 
    Model        & MNLI  & QNLI  & MRPC  & RTE & STS-B\\ \midrule
    BERT          & 84.3 & 90.6 & 88.0 & 67.8  & 88.9\\ 
    COOL(BERT)   &  84.3     & \textbf{91.3} & \textbf{91.9}      & \textbf{71.9} & \textbf{89.6} \\ \midrule
    RoBERTa & 87.6 & 92.8 & 90.2 & 78.7 & \textbf{91.2} & \\
    COOL(RoBERTa) & \textbf{87.8} & \textbf{93.0} & \textbf{93.3} & \textbf{81.3} & 90.6 
    \\
    \bottomrule
    \end{tabular}%
    }
\caption{GLUE Test results. F1 scores are reported for MRPC. Spearman correlations are reported for STS-B. Accuracy scores are reported for the other tasks.}
\vspace{-3mm}
\label{tab:nli_results}
\end{table}

\paragraph{Natural language inference.}
Natural language inference (NLI), also termed natural language understanding (NLU), targets to determine whether two input sequences entail or contradict each other ~\cite{wang-etal-2018-glue}. 
The General Language Understanding Evaluation benchmark (GLUE), composed of 9 datasets for different tasks, is widely used for NLI.
We evaluate the proposed approach on the Multi-Genre Natural Language Inference Corpus (MNLI)~\cite{williams-etal-2018-broad}, the Question-answering NLI dataset (QNLI)~\cite{rajpurkar-etal-2016-squad}, the Microsoft Research Paraphrase Corpus (MRPC)~\cite{dolan-brockett-2005-automatically}, the Recognizing Textual Entailment dataset (RTE)~\cite{rte}, and the Semantic Textual Similarity Benchmark (STS-B)~\cite{sts-b}.
The results are presented in Table~\ref{tab:nli_results}. COOL(BERT) and COOL(RoBERTa) always outperform the originals, excluding on the MNLI dataset. Moreover, COOL(BERT) and COOL(RoBERTa) achieve significant improvements on the MRPC and RTE datasets.

\paragraph{Multiple-choice question answering.}
Multiple-choice question answering (MCQA) is a type of retrieval-based question answering that selects an answer to a question by ranking the given choices. The Situations With Adversarial Generations (SWAG) dataset contains 113k sentence-pair examples that evaluate grounded commonsense inference~\cite{zellers-etal-2018-swag}. It targets to choose the most plausible continuation among four choices.
The results of the comparisons of the baselines with models with COOL are reported in Table~\ref{tab:swag_conll_imdb_results}. According to the results, COOL module is able to improve this task.

\begin{table}[t]
\centering
    \resizebox{\columnwidth}{!}{
    \begin{tabular}{lccccc}
    \toprule
    Model & SWAG & NER & POS  & SST   & IMDB\\\midrule
    BERT      & 79.7 & 90.8  & 93.4 & \textbf{92.5} &  93.4 \\
    COOL(BERT)  & \textbf{79.9} &  \textbf{91.4} & \textbf{93.7 } & 92.2  &  \textbf{93.6}\\ \midrule
    RoBERTa & 83.3 &91.9 & 92.8 & 94.8 & \textbf{91.3} \\
    COOL(RoBERTa) & \textbf{83.4} & \textbf{95.3} & \textbf{93.5} & \textbf{95.2} & 91.2 \\\bottomrule
    \end{tabular}%
    }
\caption{SWAG Test accuracy results, CoNLL-2003 NER and POS F1 results, and SST and IMDB accuracy results.}
\vspace{-3mm}
\label{tab:swag_conll_imdb_results}
\end{table}

\paragraph{Named-entity recognition and part-of-speech tagging.}
Named-entity recognition (NER) locates and classifies named entities in a sentence into pre-defined categories. It extracts, for instance, names of locations and people from the text and places them under certain categories, such as organization and person~\cite{yamada-etal-2020-luke}. Part-of-speech (POS) tagging assigns to each word in the text its part of speech, e.g., noun, verb, adjective,
and other grammatical categories such as tense~\cite{pos}. NER and POS tagging are basic NLP tasks. We evaluate the model performance on the CoNLL-2003 dataset~\cite{tjong-kim-sang-de-meulder-2003-introduction}.
The comparative results are reported in Table~\ref{tab:swag_conll_imdb_results}.
We can observe that COOL benefits the two tasks as well.

\paragraph{Sentiment analysis.}
Sentiment analysis (SA) determines whether a sentence is positive, negative, or neutral. The common datasets are the large movie review dataset (IMDB)~\cite{maas-etal-2011-learning} and the Stanford Sentiment Treebank (SST-2) from GLUE. 
The comparisons are also presented in Table~\ref{tab:swag_conll_imdb_results}.
For this task, the results are mixed, suggesting the possible lesser importance of the fine-grained element brought by involving the local context.

\subsection{Other Language}

To verify the proposed context outlooker is generalisable for other languages. We try to apply it to Malay, an important low-resource language in Southeast Asia. We evaluate the approach on Malay SQuAD 2.0 which is translated from English SQuAD 2.0 by Husein\footnote{\url{https://github.com/huseinzol05/Malay-Dataset}}. Compared with the original BERT-Bahasa\footnote{\url{https://github.com/huseinzol05/malaya}} in Table~\ref{tab:malay_results}, the proposed approach also brings improvement to the Malay language. It further proves the effectiveness and generalisability of the proposed module.
\begin{table}[t]
\centering
\small
    \begin{tabular}{lcc} \toprule
    Model        & EM & F1 \\ \midrule
    BERT-Bahasa  &  57.17 & 61.49 \\  
    COOL(BERT)   &  \textbf{58.41}& \textbf{63.24}\\   \bottomrule
    \end{tabular}%
\caption{Malay SQuAD 2.0 Dev results for Malay extractive question answering.}
\label{tab:malay_results}
\vspace{-3mm}
\end{table}


\subsection{Comparison with augmentation by a local graph}

We consider whether we can perform local augmentation by building a local graph on top of the fully connected graph processed by the self-attention mechanism. Thus, we try to build a local graph with the outputs from a backbone where only close neighbours within the predefined range are connected and adopt graph neural networks (GNNs) to update the vertex embeddings. 
In Table~\ref{tab:graph_results}, we report the preliminary results by utilising different GNNs, i.e., relational graph convolution network (RGCN)~\cite{michael2018modeling}, graph attention network (GAT)~\cite{velivckovic2017graph}, and heterogeneous graph transformer (HGT)~\cite{hgt-www-2020}. The model with RGCN underperforms the baseline as it treats all neighbours equally, which might weaken the discrimination of different neighbours. The models with GAT or HGT, considering weighted effects from different neighbours, outperform the baselines that prove the impact of local augmentation. Nevertheless, the results do not exceed the model with a context outlooker. We consider the reason is the GNNs handle the direct neighbours only. However, the context outlooker considers more constraints from indirect neighbours.

\begin{table}[!t]
\centering
    \begin{tabular}{lcc} \toprule
    Model                 & F1    & EM\\ \midrule
    BERT & 75.41 & 71.78 \\
    \quad + Local Graph (RGCN) & 74.83 & 71.52 \\ 
    \quad + Local Graph (GAT) & 75.79 & \textbf{72.90} \\
    \quad + Local Graph (HGT) & \textbf{75.96 }& 72.67 \\
    \bottomrule
    \end{tabular}
\caption{Comparisons with GNNs on SQuAD 2.0.}
\vspace{-4mm}
\label{tab:graph_results}
\end{table}

\section{Conclusion}
We presented COOL, an outlook attention mechanism for natural language processing that can be added to a transformer-based network for any natural language processing task. 

A comparative empirical performance evaluation of an implementation of COOL with different models confirms that COOL, in its \emph{Global-to-Local} mode, yields a practical performance improvement over a baseline using the original model alone for question answering. The performance and examples suggest that COOL is indeed able to acquire a more expressive representation of local context information.

We also presented empirical results for other natural language processing tasks: question generation, multiple-choice question answering, natural language inference, sentiment analysis, named-entity recognition, and part-of-speech tagging not only illustrate its portability and versatility of COOL over different models and for different tasks but also confirms the positive contribution of COOL to the effectiveness of the resolution of these tasks and other languages. 

\section*{Acknowledgments}

This research/project is supported by the National Research Foundation, Singapore under its Industry Alignment Fund – Pre-positioning (IAF-PP) Funding Initiative. Any opinions, findings and conclusions or recommendations expressed in this material are those of the author(s) and do not reflect the views of National Research Foundation, Singapore.

\bibliographystyle{named}
\bibliography{anthology_short,custom_short}

\begin{thebibliography}{}

\bibitem[\protect\citeauthoryear{Bahdanau \bgroup \em et al.\egroup
  }{2014}]{nmt2014}
Dzmitry Bahdanau, Kyunghyun Cho, and Yoshua Bengio.
\newblock Neural machine translation by jointly learning to align and
  translate.
\newblock {\em arXiv preprint arXiv:1409.0473}, 2014.

\bibitem[\protect\citeauthoryear{Banerjee and
  Lavie}{2005}]{banerjee-lavie-2005-meteor}
Satanjeev Banerjee and Alon Lavie.
\newblock {METEOR}: An automatic metric for {MT} evaluation with improved
  correlation with human judgments.
\newblock In {\em {ACL} Workshop on Intrinsic and Extrinsic Evaluation Measures
  for Machine Translation and/or Summarization}, 2005.

\bibitem[\protect\citeauthoryear{Bentivogli \bgroup \em et al.\egroup
  }{2009}]{rte}
Luisa Bentivogli, Peter Clark, Ido Dagan, and Danilo Giampiccolo.
\newblock The fifth {PASCAL} recognizing textual entailment challenge.
\newblock In {\em TAC}, 2009.

\bibitem[\protect\citeauthoryear{Cer \bgroup \em et al.\egroup }{2017}]{sts-b}
Daniel Cer, Mona Diab, Eneko Agirre, Inigo Lopez-Gazpio, and Lucia Specia.
\newblock Semantic textual similarity-multilingual and cross-lingual focused
  evaluation.
\newblock {\em arXiv preprint arXiv:1708.00055}, 2017.

\bibitem[\protect\citeauthoryear{Cheng \bgroup \em et al.\egroup
  }{2016}]{cheng2016long}
Jianpeng Cheng, Li~Dong, and Mirella Lapata.
\newblock Long short-term memory-networks for machine reading.
\newblock {\em arXiv preprint arXiv:1601.06733}, 2016.

\bibitem[\protect\citeauthoryear{Devlin \bgroup \em et al.\egroup
  }{2019}]{devlin-etal-2019-bert}
Jacob Devlin, Ming-Wei Chang, Kenton Lee, and Kristina Toutanova.
\newblock {BERT}: Pre-training of deep bidirectional transformers for language
  understanding.
\newblock In {\em the Conference of the North {A}merican Chapter of the
  Association for Computational Linguistics}, 2019.

\bibitem[\protect\citeauthoryear{Dolan and
  Brockett}{2005}]{dolan-brockett-2005-automatically}
William~B. Dolan and Chris Brockett.
\newblock Automatically constructing a corpus of sentential paraphrases.
\newblock In {\em the Workshop on Paraphrasing}, 2005.

\bibitem[\protect\citeauthoryear{Dong \bgroup \em et al.\egroup
  }{2018}]{dong2018speech}
Linhao Dong, Shuang Xu, and Bo~Xu.
\newblock Speech-transformer: A no-recurrence sequence-to-sequence model for
  speech recognition.
\newblock In {\em the Conference on Acoustics, Speech and Signal Processing},
  2018.

\bibitem[\protect\citeauthoryear{{Dosovitskiy} \bgroup \em et al.\egroup
  }{2021}]{vit}
Alexey {Dosovitskiy}, Lucas {Beyer}, Alexander {Kolesnikov}, Dirk
  {Weissenborn}, Xiaohua {Zhai}, Thomas {Unterthiner}, Mostafa {Dehghani},
  Matthias {Minderer}, Georg {Heigold}, Sylvain {Gelly}, Jakob {Uszkoreit}, and
  Neil {Houlsby}.
\newblock An image is worth 16x16 words: Transformers for image recognition at
  scale.
\newblock In {\em The Ninth International Conference on Learning
  Representations}, 2021.

\bibitem[\protect\citeauthoryear{Glorot \bgroup \em et al.\egroup
  }{2011}]{pmlr-v15-glorot11a}
Xavier Glorot, Antoine Bordes, and Yoshua Bengio.
\newblock Deep sparse rectifier neural networks.
\newblock In {\em the Conference on Artificial Intelligence and Statistics},
  2011.

\bibitem[\protect\citeauthoryear{Gulati \bgroup \em et al.\egroup
  }{2020}]{gulati20_interspeech}
Anmol Gulati, James Qin, Chung-Cheng Chiu, Niki Parmar, Yu~Zhang, Jiahui Yu,
  Wei Han, Shibo Wang, Zhengdong Zhang, Yonghui Wu, and Ruoming Pang.
\newblock Conformer: Convolution-augmented transformer for speech recognition.
\newblock In {\em Proc. Interspeech}, 2020.

\bibitem[\protect\citeauthoryear{He \bgroup \em et al.\egroup
  }{2016}]{he2016residual}
Kaiming He, Xiangyu Zhang, Shaoqing Ren, and Jian Sun.
\newblock Deep residual learning for image recognition.
\newblock In {\em IEEE Conference on Computer Vision and Pattern Recognition},
  2016.

\bibitem[\protect\citeauthoryear{He \bgroup \em et al.\egroup
  }{2021}]{He2021DeBERTaDB}
Pengcheng He, Xiaodong Liu, Jianfeng Gao, and Weizhu Chen.
\newblock {D}e{BERT}a: Decoding-enhanced {BERT} with disentangled attention.
\newblock {\em ArXiv}, abs/2006.03654, 2021.

\bibitem[\protect\citeauthoryear{Hofst{\"a}tter \bgroup \em et al.\egroup
  }{2020}]{local_for_long_text}
Sebastian Hofst{\"a}tter, Hamed Zamani, Bhaskar Mitra, Nick Craswell, and Allan
  Hanbury.
\newblock Local self-attention over long text for efficient document retrieval.
\newblock In {\em the 43rd International Conference on Research and Development
  in Information Retrieval}, 2020.

\bibitem[\protect\citeauthoryear{Hu \bgroup \em et al.\egroup
  }{2020}]{hgt-www-2020}
Ziniu Hu, Yuxiao Dong, Kuansan Wang, and Yizhou Sun.
\newblock Heterogeneous graph transformer.
\newblock In {\em Proceedings of The Web Conference}, 2020.

\bibitem[\protect\citeauthoryear{Jiang \bgroup \em et al.\egroup
  }{2020}]{jiang2020convbert}
Zi-Hang Jiang, Weihao Yu, Daquan Zhou, Yunpeng Chen, Jiashi Feng, and Shuicheng
  Yan.
\newblock Conv{BERT}: Improving {BERT} with span-based dynamic convolution.
\newblock In {\em Advances in Neural Information Processing Systems}, 2020.

\bibitem[\protect\citeauthoryear{Kalchbrenner \bgroup \em et al.\egroup
  }{2014}]{kalchbrenner-etal-2014-convolutional}
Nal Kalchbrenner, Edward Grefenstette, and Phil Blunsom.
\newblock A convolutional neural network for modelling sentences.
\newblock In {\em the Association for Computational Linguistics}, 2014.

\bibitem[\protect\citeauthoryear{Karpov \bgroup \em et al.\egroup
  }{2020}]{karpov2020transformer}
Pavel Karpov, Guillaume Godin, and Igor~V Tetko.
\newblock Transformer-{CNN}: Swiss knife for qsar modeling and interpretation.
\newblock {\em Journal of Cheminformatics}, 2020.

\bibitem[\protect\citeauthoryear{Lan \bgroup \em et al.\egroup
  }{2019}]{lan2019albert}
Zhenzhong Lan, Mingda Chen, Sebastian Goodman, Kevin Gimpel, Piyush Sharma, and
  Radu Soricut.
\newblock {Albert}: A lite {BERT} for self-supervised learning of language
  representations.
\newblock {\em arXiv preprint arXiv:1909.11942}, 2019.

\bibitem[\protect\citeauthoryear{Lin}{2004}]{lin-2004-rouge}
Chin-Yew Lin.
\newblock {ROUGE}: A package for automatic evaluation of summaries.
\newblock In {\em Text Summarization Branches Out}, 2004.

\bibitem[\protect\citeauthoryear{Liu \bgroup \em et al.\egroup
  }{2021a}]{liu2021transformer}
Yun Liu, Guolei Sun, Yu~Qiu, Le~Zhang, Ajad Chhatkuli, and Luc Van~Gool.
\newblock Transformer in convolutional neural networks.
\newblock {\em arXiv preprint arXiv:2106.03180}, 2021.

\bibitem[\protect\citeauthoryear{Liu \bgroup \em et al.\egroup }{2021b}]{swin}
Ze~Liu, Yutong Lin, Yue Cao, Han Hu, Yixuan Wei, Zheng Zhang, Stephen Lin, and
  Baining Guo.
\newblock Swin transformer: Hierarchical vision transformer using shifted
  windows.
\newblock {\em arXiv preprint arXiv:2103.14030}, 2021.

\bibitem[\protect\citeauthoryear{Loshchilov and
  Hutter}{2019}]{Loshchilov2019DecoupledWD}
Ilya Loshchilov and Frank Hutter.
\newblock Decoupled weight decay regularization.
\newblock In {\em the Conference on Learning Representations}, 2019.

\bibitem[\protect\citeauthoryear{Maas \bgroup \em et al.\egroup
  }{2011}]{maas-etal-2011-learning}
Andrew~L. Maas, Raymond~E. Daly, Peter~T. Pham, Dan Huang, Andrew~Y. Ng, and
  Christopher Potts.
\newblock Learning word vectors for sentiment analysis.
\newblock In {\em the Association for Computational Linguistics}, 2011.

\bibitem[\protect\citeauthoryear{Michael~Sejr \bgroup \em et al.\egroup
  }{2018}]{michael2018modeling}
Schlichtkrull Michael~Sejr, Kipf Thomas~N., Bloem Peter, Berg Rianne Van~Den,
  Titov Ivan, and Welling Max.
\newblock Modeling relational data with graph convolutional networks.
\newblock In {\em The Semantic Web - 15th International Conference}, 2018.

\bibitem[\protect\citeauthoryear{Papineni \bgroup \em et al.\egroup
  }{2002}]{papineni-etal-2002-bleu}
Kishore Papineni, Salim Roukos, Todd Ward, and Wei-Jing Zhu.
\newblock {BLEU}: a method for automatic evaluation of machine translation.
\newblock In {\em the Association for Computational Linguistics}, 2002.

\bibitem[\protect\citeauthoryear{Qi \bgroup \em et al.\egroup
  }{2020}]{qi-etal-2020-prophetnet}
Weizhen Qi, Yu~Yan, Yeyun Gong, Dayiheng Liu, Nan Duan, Jiusheng Chen, Ruofei
  Zhang, and Ming Zhou.
\newblock {P}rophet{N}et: Predicting future n-gram for sequence-to-{S}equence
  {P}re-training.
\newblock In {\em Findings of the Association for Computational Linguistics},
  2020.

\bibitem[\protect\citeauthoryear{Rajpurkar \bgroup \em et al.\egroup
  }{2016}]{rajpurkar-etal-2016-squad}
Pranav Rajpurkar, Jian Zhang, Konstantin Lopyrev, and Percy Liang.
\newblock {SQ}u{AD}: 100,000+ questions for machine comprehension of text.
\newblock In {\em the Conference on Empirical Methods in Natural Language
  Processing}, 2016.

\bibitem[\protect\citeauthoryear{Rajpurkar \bgroup \em et al.\egroup
  }{2018}]{rajpurkar-etal-2018-know}
Pranav Rajpurkar, Robin Jia, and Percy Liang.
\newblock Know what you don{'}t know: Unanswerable questions for {SQ}u{AD}.
\newblock In {\em the Association for Computational Linguistics}, 2018.

\bibitem[\protect\citeauthoryear{Sukhbaatar \bgroup \em et al.\egroup
  }{2019}]{sukhbaatar-etal-2019-adaptive}
Sainbayar Sukhbaatar, Edouard Grave, Piotr Bojanowski, and Armand Joulin.
\newblock Adaptive attention span in transformers.
\newblock In {\em the Association for Computational Linguistics}, 2019.

\bibitem[\protect\citeauthoryear{Sun \bgroup \em et al.\egroup
  }{2019}]{Sun2019BERT4RecSR}
Fei Sun, Jun Liu, Jian Wu, Changhua Pei, Xiao Lin, Wenwu Ou, and Peng Jiang.
\newblock {BERT}4{R}ec: Sequential recommendation with bidirectional encoder
  representations from transformer.
\newblock In {\em the Conference on Information and Knowledge Management},
  2019.

\bibitem[\protect\citeauthoryear{Tian and Lo}{2015}]{pos}
Yuan Tian and David Lo.
\newblock A comparative study on the effectiveness of part-of-speech tagging
  techniques on bug reports.
\newblock In {\em the Conference on Software Analysis, Evolution, and
  Reengineering}, 2015.

\bibitem[\protect\citeauthoryear{Tjong Kim~Sang and
  De~Meulder}{2003}]{tjong-kim-sang-de-meulder-2003-introduction}
Erik~F. Tjong Kim~Sang and Fien De~Meulder.
\newblock Introduction to the {C}o{NLL}-2003 shared task: Language-independent
  named entity recognition.
\newblock In {\em the Seventh Conference on Natural Language Learning at
  {HLT}-{NAACL}}, 2003.

\bibitem[\protect\citeauthoryear{Vaswani \bgroup \em et al.\egroup
  }{2017}]{vaswani_attention_2017}
Ashish Vaswani, Noam Shazeer, Niki Parmar, Jakob Uszkoreit, Llion Jones,
  Aidan~N Gomez, {\L}ukasz Kaiser, and Illia Polosukhin.
\newblock Attention is all you need.
\newblock In {\em Advances in neural information processing systems}, 2017.

\bibitem[\protect\citeauthoryear{Veli{\v{c}}kovi{\'c} \bgroup \em et al.\egroup
  }{2017}]{velivckovic2017graph}
Petar Veli{\v{c}}kovi{\'c}, Guillem Cucurull, Arantxa Casanova, Adriana Romero,
  Pietro Lio, and Yoshua Bengio.
\newblock Graph attention networks.
\newblock {\em arXiv preprint arXiv:1710.10903}, 2017.

\bibitem[\protect\citeauthoryear{Wang \bgroup \em et al.\egroup
  }{2018}]{wang-etal-2018-glue}
Alex Wang, Amanpreet Singh, Julian Michael, Felix Hill, Omer Levy, and Samuel
  Bowman.
\newblock {GLUE}: A multi-task benchmark and analysis platform for natural
  language understanding.
\newblock In {\em the {EMNLP} Workshop {B}lackbox{NLP}}, 2018.

\bibitem[\protect\citeauthoryear{Williams \bgroup \em et al.\egroup
  }{2018}]{williams-etal-2018-broad}
Adina Williams, Nikita Nangia, and Samuel Bowman.
\newblock A broad-coverage challenge corpus for sentence understanding through
  inference.
\newblock In {\em the Conference of the North {A}merican Chapter of the
  Association for Computational Linguistics}, 2018.

\bibitem[\protect\citeauthoryear{Wu \bgroup \em et al.\egroup
  }{2019}]{wu2019pay}
Felix Wu, Angela Fan, Alexei Baevski, Yann~N. Dauphin, and Michael Auli.
\newblock Pay less attention with lightweight and dynamic convolutions.
\newblock In {\em 7th International Conference on Learning Representations},
  2019.

\bibitem[\protect\citeauthoryear{Xiao \bgroup \em et al.\egroup
  }{2020}]{xiao2020ernie}
Dongling Xiao, Han Zhang, Yukun Li, Yu~Sun, Hao Tian, Hua Wu, and Haifeng Wang.
\newblock {ERNIE}-{GEN}: An enhanced multi-flow pre-training and fine-tuning
  framework for natural language generation.
\newblock {\em arXiv preprint arXiv:2001.11314}, 2020.

\bibitem[\protect\citeauthoryear{Xiao \bgroup \em et al.\egroup
  }{2021}]{early_conv}
Tete Xiao, Mannat Singh, Eric Mintun, Trevor Darrell, Piotr Doll{\'a}r, and
  Ross Girshick.
\newblock Early convolutions help transformers see better.
\newblock {\em arXiv preprint arXiv:2106.14881}, 2021.

\bibitem[\protect\citeauthoryear{Yamada \bgroup \em et al.\egroup
  }{2020}]{yamada-etal-2020-luke}
Ikuya Yamada, Akari Asai, Hiroyuki Shindo, Hideaki Takeda, and Yuji Matsumoto.
\newblock {LUKE}: Deep contextualized entity representations with entity-aware
  self-attention.
\newblock In {\em the Conference on Empirical Methods in Natural Language
  Processing}, 2020.

\bibitem[\protect\citeauthoryear{Yang \bgroup \em et al.\egroup
  }{2018}]{yang-etal-2018-modeling}
Baosong Yang, Zhaopeng Tu, Derek~F. Wong, Fandong Meng, Lidia~S. Chao, and Tong
  Zhang.
\newblock Modeling localness for self-attention networks.
\newblock In {\em the Conference on Empirical Methods in Natural Language
  Processing}, 2018.

\bibitem[\protect\citeauthoryear{Yuan \bgroup \em et al.\egroup }{2021}]{volo}
Li~Yuan, Qibin Hou, Zihang Jiang, Jiashi Feng, and Shuicheng Yan.
\newblock {VOLO}: Vision outlooker for visual recognition.
\newblock {\em arXiv preprint arXiv:2106.13112}, 2021.

\bibitem[\protect\citeauthoryear{Zellers \bgroup \em et al.\egroup
  }{2018}]{zellers-etal-2018-swag}
Rowan Zellers, Yonatan Bisk, Roy Schwartz, and Yejin Choi.
\newblock {SWAG}: A large-scale adversarial dataset for grounded commonsense
  inference.
\newblock In {\em the Conference on Empirical Methods in Natural Language
  Processing}, 2018.

\bibitem[\protect\citeauthoryear{Zhang \bgroup \em et al.\egroup
  }{2020}]{zhang2020dynamic}
Zhebin Zhang, Sai Wu, Gang Chen, and Dawei Jiang.
\newblock Self-attention and dynamic convolution hybrid model for neural
  machine translation.
\newblock In {\em IEEE International Conference on Knowledge Graph (ICKG)},
  2020.

\bibitem[\protect\citeauthoryear{Zhuang \bgroup \em et al.\egroup
  }{2021}]{zhuang-etal-2021-robustly}
Liu Zhuang, Lin Wayne, Shi Ya, and Zhao Jun.
\newblock A robustly optimized {BERT} pre-training approach with post-training.
\newblock In {\em the 20th Chinese National Conference on Computational
  Linguistics}, 2021.

\end{thebibliography}

\end{document}